# Effective Synthetic Image Detection via Noise Residual Clustering

*Caihui Yan*[1], *Gang Cao*[1,*], *Huawei Tian*[2], *Zhen Li*[1], *Yuhang Zhai*[1]
[1] School of Computer and Cyber Sciences, Communication University of China, Beijing 100024, China
[2] People's Public Security University of China, Beijing 100038, China

***Abstract*—The rapid advancement of generative artificial intelligence (AI) has made synthetic images remarkably realistic, posing security threats such as misinformation and fraud. It is significant to detect the synthetic image in the manner of passive and blind image authentication. Most existing detectors rely on supervised training with large labeled datasets, leading to high costs and degraded performance on unknown generative models. To attenuate such deficiencies, we propose a training-free detection method. Specifically, noise residual fingerprints are first extracted by a simple yet effective pre-trained Noiseprint++ model. Then multi-scale features are further extracted from such residual by a frozen Vision Transformer (ViT), followed by adaptive weighted fusion. Only a few real image samples are used needed to initialize the clustering centers for unsupervised K-Means, distinguishing real and synthetic images without training. Extensive evaluations on four benchmark datasets show that our proposed scheme achieves an average accuracy of 82.2%, outperforming the state-of-the-art detectors on generalization ability. Superior performance is gained on the popular diffusion type of synthetic images, and the effectiveness of each module is validated by ablation studies. Source code will be publicly available at https://github.com/multimediaFor/NoiseCluSID.**

## I. INTRODUCTION

The rapid advancement of generative artificial intelligence (AI) techniques has made AI-generated images and videos visually indistinguishable from real photographs, raising serious concerns on potential misuse of such synthetic images. Generative Adversarial Networks (GANs), such as ProGAN [1], StyleGAN [2], and StarGAN [3], can synthesize highly realistic facial images. More recently, diffusion models, including Stable Diffusion [4], DALL·E 2 [5] and Midjourney [6], have achieved superior generation quality and become mainstream. The resulting AI-generated images may be maliciously exploited to spread disinformation, manipulate public opinion, and undermine the security of digital content. Therefore, it is significant to develop generalizable and robust blind detectors capable of distinguishing AI-generated images/videos from real ones [7-25].

[*] Corresponding author: Gang Cao(gangcao@cuc.edu.cn)

Existing synthetic image detection methods can be broadly categorized into supervised and training-free types. Supervised approaches train classifiers on large labeled datasets to distinguish real from synthetic images. Early efforts established strong baselines by fine-tuning off-the-shelf networks (Marra et al. [7]), and were later improved by architectural modifications such as inserting residual layers and removing early downsampling (Gragnaniello et al. [8]). Durall et al. [9] and Frank et al. [10] showed that CNN-generated images contain detectable frequency artifacts, while Corvi et al. [11] extended detection to latent diffusion models. A notable milestone is Wang et al. [12], who designed a ResNet50-based detector with aggressive data augmentation that significantly boosted generalization and became a widely adopted benchmark. Recent works diverge into several directions. Leveraging vision-language models, UnivFD [13] trains a linear classifier on frozen CLIP [14] features, achieving promising cross-model generalization. CLIP-Bar [15] and C2P-CLIP [16] further enhance the exploitation of CLIP representations for this task. Reconstruction-error-based methods DIRE [17] and DRCT [18] detect diffusion-generated images by measuring reconstruction discrepancies or employing contrastive training with a pretrained diffusion model. LGrad [19] learns generalized artifact representations from gradient information. More recently, AIDE [20] fuses high-level semantic embeddings with low-level patch artifacts, and SAFE [21] improves generalization by preventing artifact degradation and overfitting through a lightweight image transformation pipeline. Despite their effectiveness on known generators, supervised methods often suffer from poor cross-model generalization and high retraining costs.

To address these limitations, training-free detection methods have been introduced to avoid the training process altogether. The AEROB scheme [22] takes advantage of the autoencoder inside latent diffusion models. It passes an image through the encoder and decoder of a pretrained diffusion model, and then uses the discrepancy between the original and the reconstructed image as evidence for detection; real images tend to produce different reconstruction patterns compared with synthetic ones. The method requires no fine-tuning or additional data. The RIGID scheme [23] starts from the observation that real images are more resistant to slight noise perturbations than AI-generated images. It adds small-magnitude Gaussian noise to an image and examines the consistency of deep features before and after the perturbation. Real images typically maintain a stable representation, while synthetic ones exhibit a noticeable change. This perturbation-based comparison works without any training step and has shown competitive performance. Other training-free approaches, such as the few-shot learning method FTNet [24], have reported promising results. Nonetheless, detecting universal forensic traces without involving a training process remains an open problem, since current solutions often depend on priors tied to specific model architectures and may not generalize to unseen generation methods.

In this paper, we propose a novel training-free detection framework that distinguishes AI-generated images from real ones. Existing training-free methods such as AEROB [22] and RIGID [23] rely on a single type of discriminative clue derived from a single pretrained model, namely the reconstruction error of an autoencoder or the feature sensitivity to noise perturbation. In contrast, our approach captures multi-scale forensic traces by extracting hierarchical features from multiple layers of a Vision Transformer and adaptively fusing them according to their discriminative power ability. Moreover, we introduce a real-image prior guided clustering initialization using only a small set of authentic samples, which ensures stable and semantically correct clustering without any training. Extensive experimental results on multiple public datasets verify that the proposed method achieves high detection accuracy and strong cross-generator generalization, outperforming existing state-of-the-art approaches.

The remainder of this paper is organized as follows. Section II details synthetic image detection algorithm. Section III presents experimental results and analysis. Section IV concludes the paper.

## II. PROPOSED SCHEME

In this section, the proposed training-free detection scheme exploits intrinsic noise-level discrepancies between real and AI-generated images. An overview of the scheme is illustrated in Fig. 1. It consists of four core modules: image pre-processing, noise residual extraction, multi-scale feature extraction and adaptive fusion, and prior-guided unsupervised clustering. Given an input image, a pre-trained Noiseprint++ model [26] is first used to extract a noise residual map, which suppresses image content and highlights generation artifacts. From such a noise map, two types of forensic features are extracted in parallel: the deep feature captured by a Vision Transformer at multiple layers and adaptively fused, and handcrafted statistical features computed over different sliding windows. Such two complementary representations are concatenated to yield the final feature vector, which combines physical-level noise statistics with semantic-level representations. Finally, a small set of real reference images is used to initialize the clustering centers, and K-Means is performed to distinguish real and synthetic images without model training.

### *A. Noise Residual Extraction*

Various noise residual extraction techniques have been explored in prior works [26-29], and we adopt Noiseprint++ [26] due to its sensitivity to generation artifacts. Real images inherently contain sensor-related noise patterns introduced during the acquisition process, whereas AI-generated images lack suchphysical traces and instead exhibit generation-specific artifacts. To capture such discrepancies, a pre-trained Noiseprint++ model is adopted to extract the noise residual fingerprint. Let the input images be denoted as $I_i \in \mathbb{R}^{H\times W\times 3}$, where $i = 1, 2, ..., N$ and $N$ is the number of image samples. The Noiseprint++

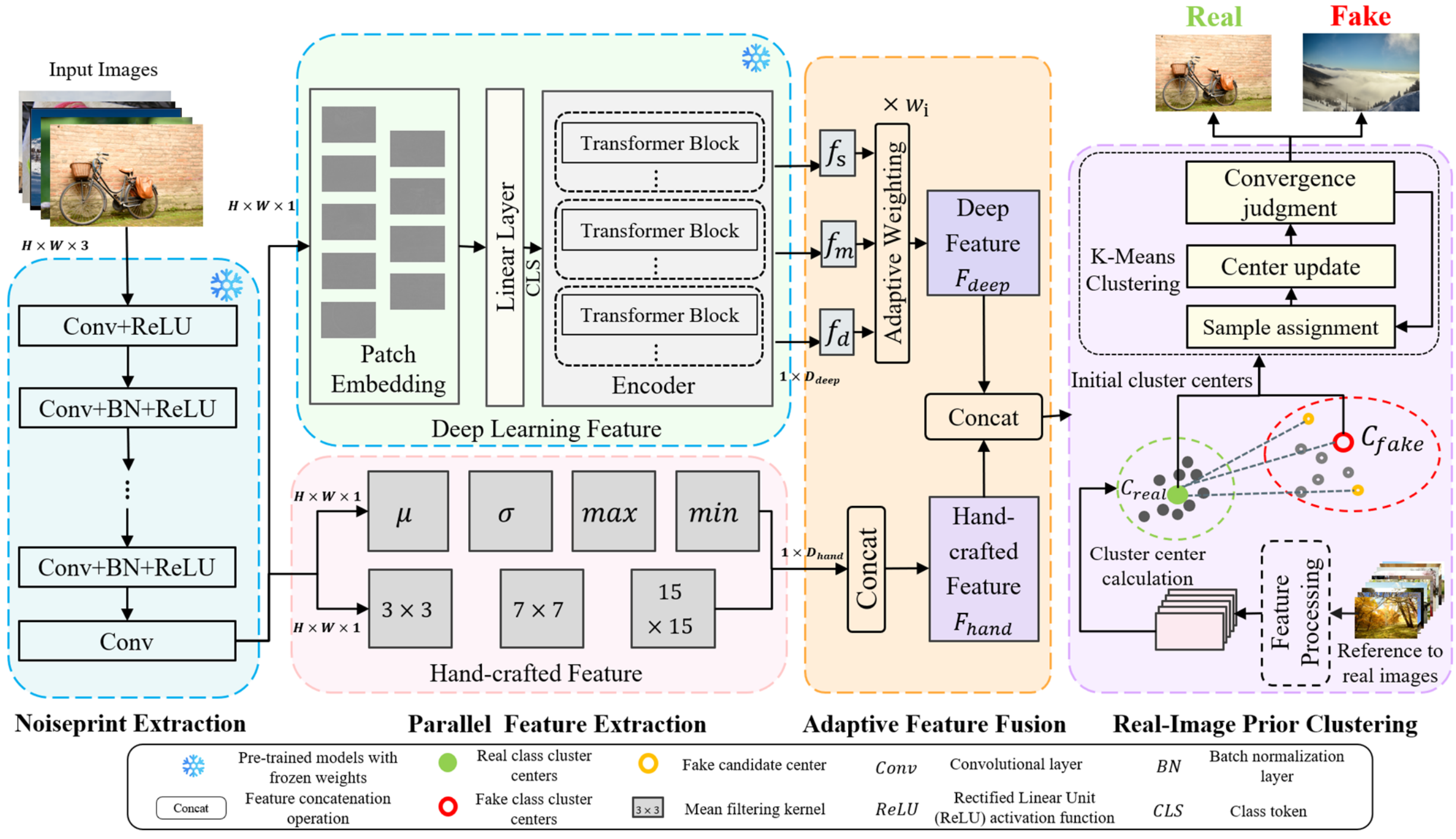

Fig. 1. Overview of the proposed training-free synthetic image detection scheme. The feature processing part of the "Real-Image Prior Clustering" module is the feature extraction and fusion shown in the three preceding modules.

model is built upon a 15-layer DnCNN architecture and trained with contrastive learning to encode both camera-internal and editing-related artifacts [26]. For each normalized input $I_i$ ( pixel values in $[0,1]$ ), the DnCNN outputs a 64-channel feature tensor $T_i \in \mathbb{R}^{H\times W\times 64}$, which is then reduced to a single-channel noise residual map $R_i \in \mathbb{R}^{H\times W}$ via channel-wise averaging:

$$R_i = \frac{1}{64}\sum_{c=1}^{64} T_i^{(c)} \tag{1}$$

The entire extraction process is performed without gradient computation, as all parameters are frozen.

### B. *Multi-Scale Feature Extraction and Fusion*

The resulting noise residual map is fed into a frozen standard Vision Transformer (ViT-B/16) [30] with ImageNet-21k pre-trained weights to extract hierarchical deep features, as illustrated in Fig. 2. Shallow transformer layers primarily respond to local textures and edges, while deeper layers capture global structural patterns. Since generation artifacts manifest at multiple scales, we extract the [CLS] token outputs from the 4th, 8th and 12th layers. Such outputs are denoted as $f_s$, $f_m$ and $f_d$, corresponding to the shallow, middle, deep features, repectively, each a 768-dimensional feature vector.

To exploit the varying discriminative power of different layers, an adaptive weighting mechanism based on the silhouette score is proposed. For each layer, the feature matrix is first Z-score standardized.

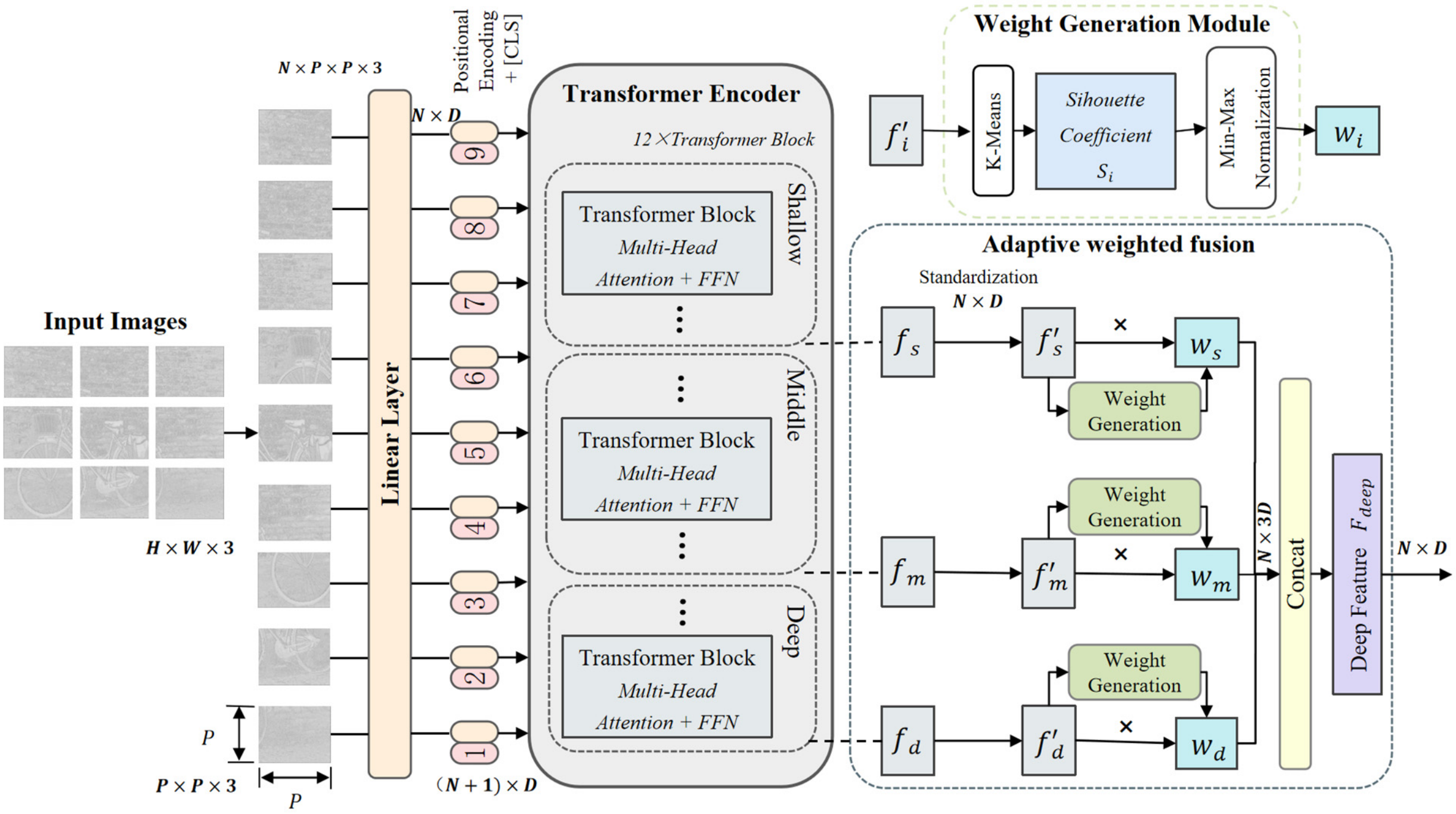

Fig. 2. Vision Transformer's multi-layer feature extraction and adaptive fusion architecture, where shallow, middle, and deep layers are weighted by their discriminative power.

K-Means clustering (*K=2*) is then applied, and the silhouette score $S_i$ is computed to quantify the cluster separation quality:

$$S_i = \frac{1}{N}\sum_{i=1}^{N}\frac{b(i)-a(i)}{\max(a(i),b(i))} \tag{2}$$

where $a(i)$ is the mean intra-cluster distance and $b(i)$ is the mean nearest-cluster distance for the *i*-th sample. The silhouette score ranges in [-1,1], with higher values indicating better cluster separation. The fusion weight $w_i$ for each layer is derived by normalizing $S_i$ across the three layers:

$$w_i = \frac{S_i - S_{\min}}{S_{\max} - S_{\min} + \varepsilon} \tag{3}$$

where $\varepsilon$ is a small constant to prevent division by zero. This normalization maps the weights to [0,1], assigning the largest weight to the most discriminative layer. The fused deep representation is obtained by weighted concatenation:

$$F_{deep} = Concat(w_s \cdot f_s^{'}, w_m \cdot f_m^{'}, w_d \cdot f_d^{'}) \tag{4}$$

where $f_s^{'}$, $f_m^{'}$, $f_d^{'}$ denote the Z-score standardized versions of $f_s$, $f_m$, $f_d$, respectively. Such a dynamic weighting strategy enables the model to automatically emphasize the most discriminative scale for each batch.

In parallel, handcrafted statistical features are also computed from the same noise residual map to provide complementary low-level cues that are robust to content variations. Such a design is inspired by the multi-view feature fusion methodology used in the local forgey localization task [31]. Specifically, for each noise residual map $R_i$, we compute four global statistics: the mean $\mu$, standard deviation $\sigma$, maximum and minimum. To further capture local noise patterns at multiple scales, we apply average pooling with kernel sizes of 3×3, 7×7, and 15×15, and compute the mean value within each pooling region, then average over all regions to obtain a single feature per kernel size. This yields three additional local mean features. In total, seven handcrafted features are extracted: $\mu$, $\sigma$, *max*, *min*, and the averaged local means for the three kernel sizes. The resulting handcrafted feature vector is denoted as $F_{hand} \in \mathbb{R}^7$. Finally, $F_{hand}$ is concatenated with the deep fused feature $F_{deep}$ to form the final feature vector $F = [F_{deep}; F_{hand}]$. By combining low-level noise statistics with high-level semantic representations, this fused feature space characterizes both the physical noise properties and the structural inconsistencies between real and generated images.

### *C. Real Image Prior-Guided Clustering*

Standard K-Means clustering suffers from sensitivity to random initialization, which may lead to unstable cluster assignments or semantic misalignment between clusters. To address this issue, a prior-guided initialization strategy is designed using only a small set of real reference images. This reference set is constructed offline before system deployment and remains fixed for all subsequent inference tasks, requiring no annotations from the test data.

The initialization process is illustrated in Fig. 3. Let the reference set contain *M* real images, and denote the feature vector of the $i$-th reference image as $f_{ref,i} \in \mathbb{R}^d$, extracted through the procedure described in Section II-A. The real-class center is computed as the mean of all reference features:

$$C_{real} = \frac{1}{M} \sum_{i=1}^{M} f_{ref,i} \tag{5}$$

For a given test batch, the Euclidean distances from all samples to $C_{real}$ are first computed. To robustly determine the fake-class center, a multi-candidate optimization strategy is adopted. Specifically, several samples farthest from $C_{real}$ are selected as candidates. For each candidate $p$, a tentative K-Means partition is performed with initial centers $\{C_{real}, p\}$, and the clustering inertia $J$ is computed as:

$$J = \sum_{k=1}^{2} \sum_{x \in C_k} \left\| x - \mu_k \right\|^2 \tag{6}$$

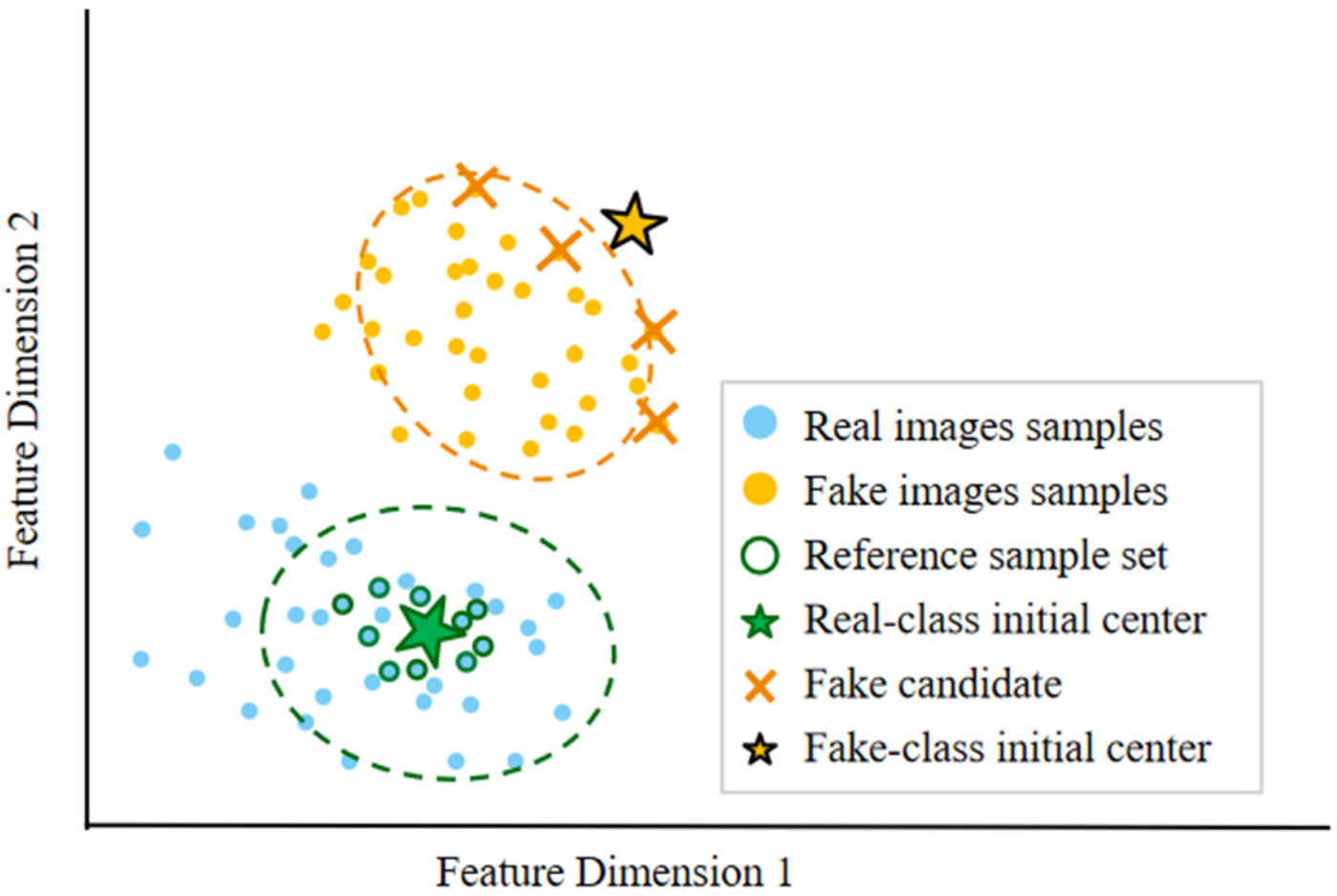

Fig. 3. Illustration of real image prior-guided clustering center initialization, where the real center is fixed and the fake center is optimized via multi-candidate selection.

The candidate that yields the minimum inertia is selected as the fake-class initial center $C_{fake}$. This strategy effectively filters out outlier interference and ensures that the initial clusters are compact. Finally, K-Means clustering is executed with $\{C_{real}, C_{fake}\}$ as fixed initial centers. During iterative updates, $C_{real}$ is kept frozen to maintain correct semantic alignment, while $C_{real}$ is updated according to the current cluster assignment. After convergence, the cluster associated with $C_{real}$ is labeled as real, and the other as generated. The entire process requires no model training and relies solely on a small number of real reference images for prior guidance.

## III. EXPERIMENTAL RESULTS AND DISCUSSION

In this section, extensive experiments are conducted to evaluate the performance of the proposed training-free detection scheme. The experimental setup including datasets, evaluation metrics, and the baseline methods used for comparison is first introduced. Performance comparison with the state-of-the-art methods is then presented, followed by ablation studies for validating the contribution of each core module. Qualitative visualization is also provided.

### *A. Experimental Setup*

**Datasets and Metrics.** To evaluate the generalization ability of the proposed scheme, we employ four public datasets, i.e., GenImage [32], Synthbuster [33], AIGCDetect (AIGCD) [34] and Chameleon [20]. These datasets cover a diverse range of generative architectures, including both GANs (e.g., ProGAN [1], StyleGAN [2], StyleGAN2 [35], BigGAN [36]) and diffusion models (e.g., Stable Diffusion [4], Midjourney [6], DALL·E 2 [5]), with spatial resolutions ranging from 256×256 to 3840×2160 pixels. GenImage and Synthbuster are primarily composed of diffusion model outputs, while Chameleon includes post-processed images. A detailed summary of these datasets is provided in Table 1. Following prior works [13, 22], we adopt Accuracy (ACC) and Average Precision (AP) as evaluation metrics, with generated images treated as the positive class.

Table 1. Summary of the four test datasets.

| Dataset | Real/Fake | Source of Real | Generator Type | Generators |
|---|---|---|---|---|
| GenImage [32] | 50K/50K | ImageNet | Diffusion + GAN | 8 |
| Synthbuster [33] | 1K/9K | RAISE | Diffusion | 9 |
| AIGCDetect [34] | 74.3K/74.3K | LSUN, ImageNet, etc. | Diffusion + GAN | 16 |
| Chameleon [20] | 14.9K/11.2K | Internet | Unknown | Unknown |

**Baseline Methods.** To evaluate the performance advantage of the proposed scheme, ten representative detection algorithms are compared as baselines, spanning both supervised and training-free ones.

The selected supervised baselines include Lgrad [19], DIRE [17], UnivFD [13], DRCT [18], CLIP-Bar [15], AIDE [20], C2P-CLIP [16], and SAFE [21]. For training-free approaches, we compare with AEROB [22] and RIGID [23]. All baselines are evaluated with their default or recommended settings, and their detailed implementations are described in the corresponding references.

**Implementation Details.** The proposed algorithm is implemented in PyTorch. All experiments run on a PC with 64GB RAM and one NVIDIA RTX 3090 GPU. The pre-trained Noiseprint++ and ViT-B/16 sub-networks are frozen throughout the entire process. The test images are first resized to 224x224 pixels and normalized to [0, 1] before being fed into the detector. The number of reference real images is set as $M$=10, and the number of farthest candidates for multi-candidate optimization is set as $K$=5. No model training is performed. All clustering operations are performed using the K-Means implementation from scikit-learn. The only non-default parameter is the use of fixed initial centers; all other parameters remain at their default values.

*B. Generalization Performance Evaluation*

In this subsection, the generalization ability of the proposed scheme across diverse test datasets is evaluated. Tables 2 reports the ACC and AP of all compared methods on four datasets, with the average across datasets reported in the last column to reflect overall generalization. The proposed method achieves the highest average ACC of 85.9% and average AP of 87.5%, outperforming all ten baseline methods.

The proposed method exhibits a pronounced advantage on diffusion model-dominated datasets. On Synthbuster, ACC reaches 97.7% and on GenImage ACC reaches 97.5%, both substantially higher than the second-best results. This can be attributed to our proposed feature extraction and fusion framework (NoiseCluSID), which combines noise residuals with multi-scale ViT features and adaptive weighting.

Table 2. Detection accuracy (ACC) and average precision (AP) comparison with other methods on different datasets. The best result in each column is highlighted in bold and the second-best is underlined.

| Method | Synthbuster ACC/AP | GenImage ACC/AP | AIGCD ACC/AP | Chameleon ACC/AP | AVG ACC/AP |
|---|---|---|---|---|---|
| Lgrad [19] | 0.443/0.432 | 0.519/0.522 | 0.502/0.522 | 0.485/0.380 | 0.487/0.464 |
| DIRE [17] | 0.476/0.479 | 0.619/0.544 | 0.537/0.554 | 0.467/0.436 | 0.524/0.503 |
| UnivFD [13] | 0.616/0.604 | 0.692/0.911 | 0.784/0.911 | 0.573/0.468 | 0.666/0.723 |
| DRCT [18] | 0.730/0.770 | 0.835/**1.000** | 0.707/0.800 | 0.699/0.708 | 0.742/0.819 |
| CLIP-Bar [15] | 0.739/0.912 | 0.818/0.941 | 0.775/0.894 | 0.554/0.461 | 0.721/0.802 |
| AIDE [20] | 0.661/0.644 | 0.862/**1.000** | 0.819/0.931 | 0.642/0.664 | 0.746/0.809 |
| C2P-CLIP [16] | 0.461/0.581 | 0.966/0.998 | 0.835/0.899 | 0.541/0.536 | 0.700/0.753 |
| SAFE [21] | 0.527/0.574 | 0.955/0.998 | **0.944/0.980** | 0.593/0.481 | 0.754/0.758 |
| AEROB [22] | 0.636/0.592 | 0.693/0.689 | 0.579/0.584 | 0.477/0.535 | 0.596/0.600 |
| RIGID [23] | 0.593/0.611 | 0.865/0.875 | 0.759/0.762 | 0.521/**0.722** | 0.684/0.742 |
| Ours | **0.977/0.981** | **0.975**/0.977 | 0.801/0.822 | **0.681**/0.719 | **0.859/0.875** |

Such a design is particularly effective at capturing the over-smoothing artifacts inherent in the iterative denoising process of diffusion models, while the multi-scale fusion further enhances the discriminability of these traces. Supervised methods such as SAFE [21] and C2P-CLIP [16] achieve noticeably lower scores on these two datasets, indicating that their learned features, though effective on seen generators, do not transfer as effectively when training data coverage is limited.

On the Chameleon dataset consisting of post-processed images, our method achieves the best ACC, i.e., 68.1%, with a smaller performance drop relative to its results on clean datasets compared with most baselines. The above results suggest that the proposed feature representation exhibits a certain degree of robustness under common post-processing operations such as compression and scaling.

Comparing with the prior training-free methods, the proposed method significantly outperforms AEROB (average ACC 59.6%, AP 60%) and RIGID (average ACC 68.4%, AP 74.2%). AEROB relies solely on autoencoder reconstruction error, which is effective only when the test image is generated by a model sharing the same encoder architecture. RIGID analyzes perturbation sensitivity at a single scale, whereas the proposed method fuses features from multiple transformer layers, capturing forensic traces at varying granularities. This multi-scale strategy proves to be a key factor behind the performance gain.

Tables 3 and 4 present fine-grained detection accuracy (ACC) comparisons on the Synthbuster and GenImage datasets, respectively. On Synthbuster (Table 4), the proposed method achieves near□perfect detection on most diffusion□based models, including Midjourney, the Stable Diffusion series, and GLIDE, substantially outperforming all baseline methods. On DALL·E 2 and DALL·E 3, the ACC remains above 95%. The only noticeable drop occurs on Firefly (74.7%), which is still better than the second-best result. This degradation may be attributed to additional post-processing in the Firefly pipeline that partially masks the forensic traces exploited by our noise residual features.

Table 3. Fine-grained detection accuracy (ACC) comparison on each generative model subset within Synthbuster. The best result in each column is highlighted in bold and the second-best result is underlined.

| Model | D·E2 | D·E3 | Fire. | Glide | Mid | SD1.3 | SD1.4 | SD2 | SDXL | AVG |
|---|---|---|---|---|---|---|---|---|---|---|
| Lgrad [19] | 0.482 | 0.408 | 0.419 | 0.615 | 0.449 | 0.403 | 0.403 | 0.407 | 0.405 | 0.443 |
| DIRE [17] | 0.536 | 0.364 | 0.515 | 0.659 | 0.519 | 0.351 | 0.346 | 0.501 | 0.496 | 0.476 |
| UnivFD [13] | 0.527 | 0.549 | 0.507 | 0.622 | 0.583 | 0.578 | 0.995 | 0.562 | 0.625 | 0.616 |
| DRCT [18] | 0.476 | 0.534 | 0.481 | 0.602 | 0.893 | 0.941 | 0.941 | 0.908 | 0.794 | 0.729 |
| CLIP-Bar [15] | 0.681 | 0.742 | 0.646 | 0.972 | 0.629 | 0.788 | 0.785 | 0.714 | 0.689 | 0.738 |
| AIDE [20] | 0.472 | 0.559 | 0.412 | 0.786 | 0.761 | 0.752 | 0.939 | 0.647 | 0.621 | 0.661 |
| C2P-CLIP [16] | 0.496 | 0.498 | 0.173 | 0.499 | 0.499 | 0.501 | 0.501 | 0.483 | 0.500 | 0.461 |
| SAFE [21] | 0.643 | 0.162 | 0.160 | 0.585 | 0.631 | 0.658 | 0.654 | 0.594 | 0.658 | 0.527 |
| AEROB [22] | 0.499 | 0.499 | 0.499 | 0.961 | 0.589 | 0.797 | 0.799 | 0.500 | 0.579 | 0.636 |
| RIGID [23] | 0.591 | 0.506 | 0.548 | 0.804 | 0.585 | 0.507 | 0.500 | 0.601 | 0.694 | 0.593 |
| Ours | **0.954** | **0.978** | **0.747** | **0.994** | **1.000** | **0.999** | **0.996** | **0.971** | **0.936** | **0.953** |

Table 4. Fine-grained detection accuracy (ACC) comparison on each generative model subset within GenImage. The best result in each column is highlighted in bold and the second-best result is underlined.

| Model | ADM | BigGAN | Mid | VQDM | Glide | SD1.4 | SD1.5 | Wukong | AVG |
|---|---|---|---|---|---|---|---|---|---|
| Lgrad [19] | 0.534 | 0.506 | 0.568 | 0.514 | 0.601 | 0.471 | 0.467 | 0.493 | 0.519 |
| DIRE [17] | 0.751 | 0.596 | 0.876 | 0.549 | 0.629 | 0.513 | 0.518 | 0.521 | 0.619 |
| UnivFD [13] | 0.625 | 0.901 | 0.551 | 0.853 | 0.625 | 0.636 | 0.634 | 0.709 | 0.692 |
| DRCT [18] | 0.663 | 0.593 | 0.943 | 0.767 | 0.732 | 0.992 | 0.991 | **0.992** | 0.834 |
| CLIP-Bar [15] | 0.665 | 0.912 | 0.681 | 0.827 | 0.959 | 0.854 | 0.857 | 0.785 | 0.817 |
| AIDE [20] | 0.733 | 0.668 | 0.793 | 0.802 | 0.917 | **0.997** | **0.997** | 0.988 | 0.862 |
| C2P-CLIP [16] | 0.902 | 0.943 | **0.975** | **0.963** | **0.978** | 0.989 | 0.987 | 0.986 | **0.965** |
| SAFE [21] | 0.820 | 0.977 | 0.952 | 0.962 | 0.962 | 0.993 | 0.992 | 0.982 | 0.955 |
| AEROB [22] | 0.541 | 0.514 | 0.674 | 0.751 | 0.854 | 0.668 | 0.674 | 0.864 | 0.692 |
| RIGID [23] | 0.821 | 0.875 | 0.895 | 0.796 | 0.853 | 0.871 | 0.881 | 0.923 | 0.864 |
| Ours | **0.964** | **0.994** | 0.969 | 0.607 | 0.968 | 0.938 | 0.943 | 0.944 | 0.916 |

Table 5. Ablation study of the proposed components. Accuracy (ACC) is reported. "w/o" denotes "without" the corresponding module.

| Variant | Synthbuster | GenImage | AIGCD | Chameleon | AVG |
|---|---|---|---|---|---|
| Level Ⅰ: Feature Extraction | | | | | |
| w/o Noiseprint (Raw RGB) | 0.850 | 0.861 | 0.705 | 0.623 | 0.760 |
| w/o Multi-scale (Last layer only) | 0.951 | 0.965 | 0.765 | 0.682 | 0.841 |
| w/o Handcrafted (ViT only) | 0.958 | 0.970 | 0.775 | **0.688** | 0.848 |
| w/o Adaptive Fusion (Direct concat) | 0.965 | 0.972 | 0.790 | 0.685 | 0.853 |
| Level Ⅱ: Clustering | | | | | |
| w/o Prior Guidance (Std. K-Means) | 0.943 | 0.966 | 0.793 | 0.681 | 0.846 |
| Ours (Full) | **0.977** | **0.975** | **0.801** | 0.681 | **0.858** |

On GenImage (Table 4), the proposed method achieves an average ACC of 91.6% across eight tested models, ranking third among compared methods and first among training□free methods. It attains the highest ACC on BigGAN and ranks in the top two on most other remaining subsets. The variation across subsets reflects the diversity of generation artifacts; notably, the noise residual features are particularly effective for diffusion-based models.

### C. *Ablation Studies*

To systematically evaluate the contribution of each proposed module, we conduct ablation experiments following the "w/o" (without) paradigm. Starting from the full configuration, we sequentially remove or replace key components while keeping all other settings fixed. The results on the four benchmark datasets are summarized in Table 5. The variants are organized according to the feature extraction pipeline (Level I) and the clustering strategy (Level II).

Noise residual extraction serves as the foundation of our detection framework. The first variant removes the Noiseprint pre-processing and feeds raw RGB images directly into the feature extractor (w/o Noiseprint). This leads to a drastic performance drop to an average accuracy of 76.0%, which is 9.8 percentage points lower than the full model. This significant degradation validates that the noise residual fingerprints extracted by Noiseprint are essential for suppressing semantic image content and amplifying the subtle artifacts introduced by generative models.

Multi-scale feature extraction plays a critical role in capturing generation traces at different depths. When we abandon the multi-layer feature extraction and only utilize the last-layer [CLS] token of ViT (w/o Multi-scale), the average accuracy declines from 85.8% to 84.1%. This 1.7 percentage point drop is the largest among all feature-related variants, indicating that generation traces manifest at various depths of the transformer. Incorporating hierarchical features from shallow, middle, and deep layers provides complementary discriminative cues that are critical for robust detection.

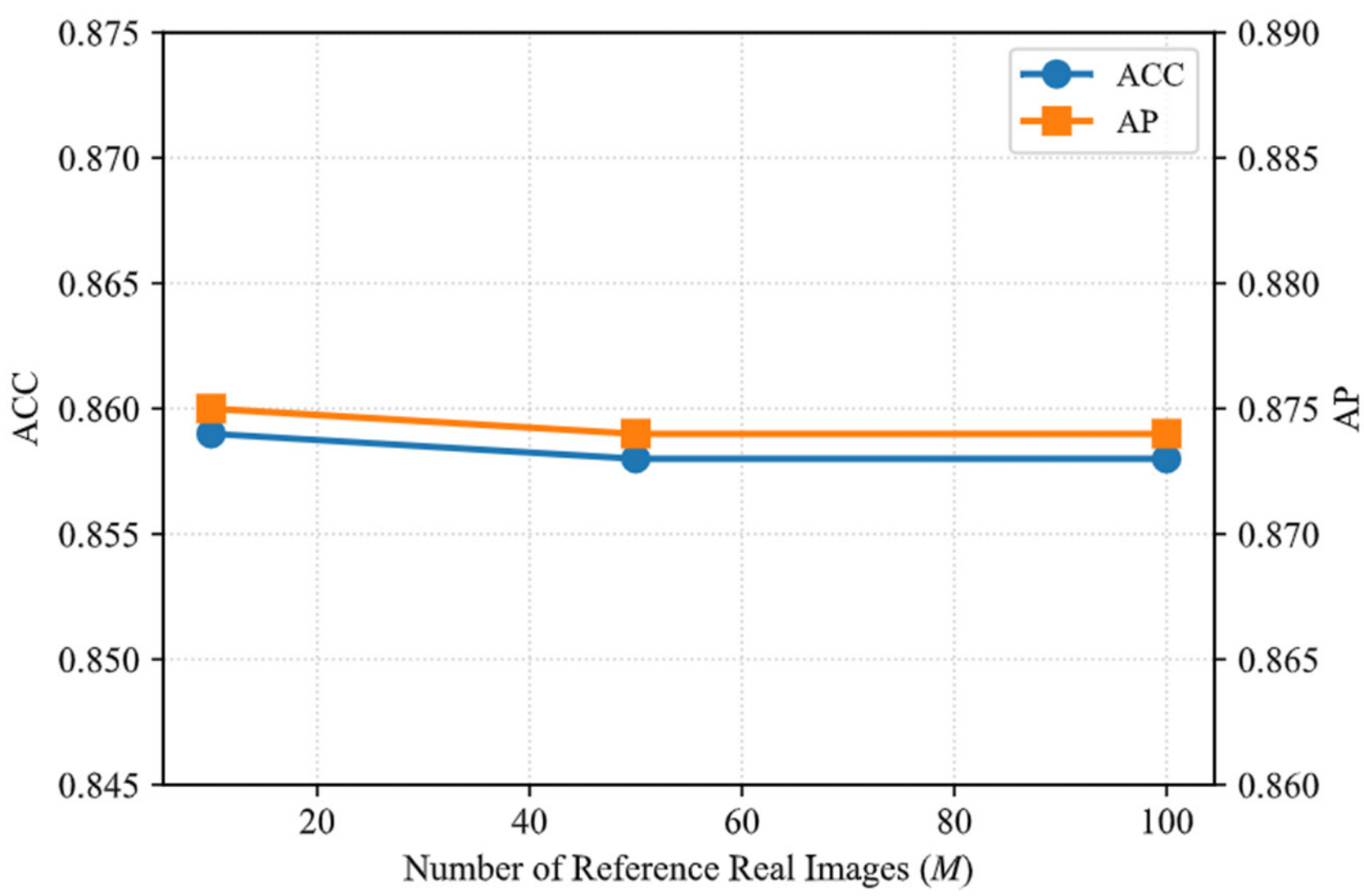

Fig. 4. Impact of reference set size (M) on detection performance: ACC and AP remain stable as M increases from 10 to 100.

Handcrafted statistical features provide complementary information to the deep semantic representations. Excluding the seven handcrafted statistical features (w/o Handcrafted) results in an average accuracy of 84.8%, which is 1.0 percentage point lower than the complete version. Although the handcrafted features contribute a relatively modest gain compared to the deep semantic features, they consistently improve detection across all datasets, especially on AIGCDetect. This demonstrates that low-level noise statistics provide beneficial complementary information to the high-level ViT representations.

Adaptive weighted fusion further refines the combined feature representation. Replacing the proposed silhouette-score-based adaptive weighting with simple direct concatenation (w/o Adaptive Fusion) yields an average accuracy of 85.3%. The full model outperforms this variant by 0.5 percentage points. While this improvement is numerically smaller than the other modules, it is consistently observed across all four datasets without introducing any trainable parameters, confirming that assigning larger weights to transformer layers with higher discriminative power effectively enhances the fused feature representation in a lightweight manner.

Prior-guided clustering effectively anchors the real class center and prevents semantic misalignment. When we substitute the prior-guided initialization with standard K-Means random initialization (w/o Prior Guidance), the average accuracy drops to 84.6%, compared to 85.8% for the full model. This 1.2 percentage point improvement is particularly noticeable on the Synthbuster and GenImage datasets, where the gap exceeds 3 percentage points. The reason is that the fixed real-class center, derived from only 10 reference real images, effectively anchors the correct semantic cluster during K-Means iterations, preventing the misalignment issues often caused by random seeds.

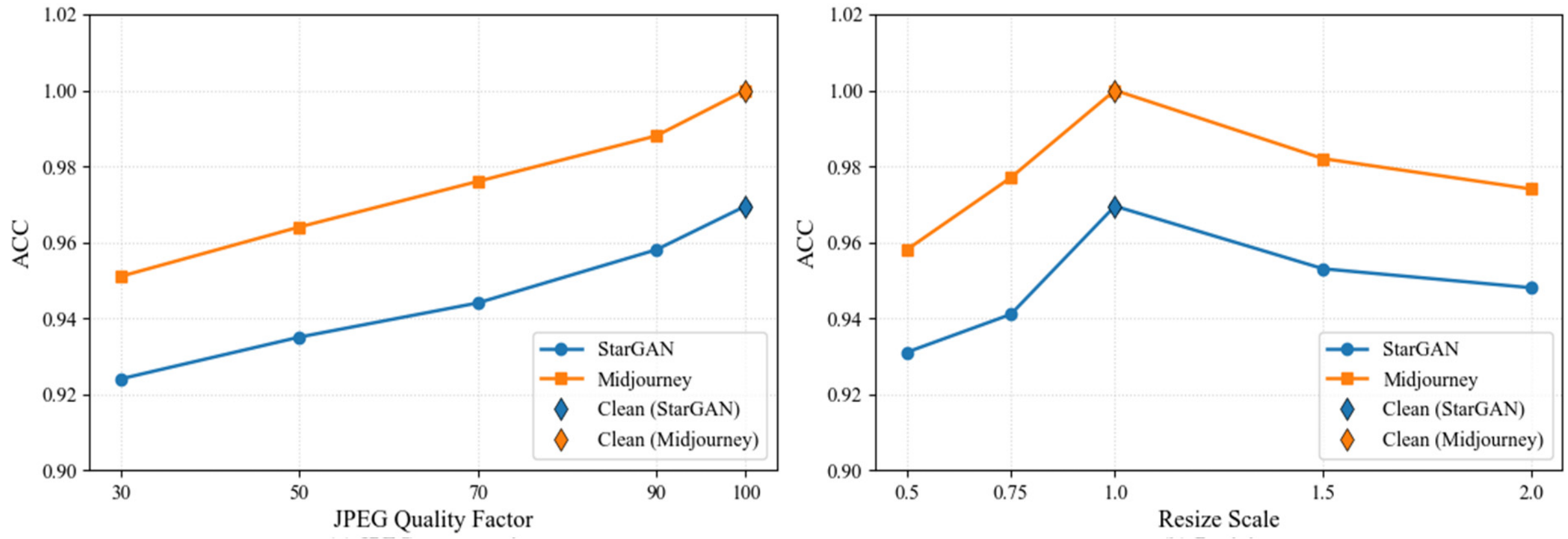

Fig. 5. Robustness evaluation of the proposed method. *Left*: against JPEG compression with different quality factors, *Right*: against resizing with different scale factors. The diamond markers denote the clean (uncompressed) baselines corresponding to QF=100 and scale=1.0, respectively.

The proposed scheme is robust to the number of reference real images. Fig. 4 reports the results for *M*=10, 50, and 100, where both ACC and AP exhibit stable performance with a maximum variation below 0.5%. In practice, we also empirically observe that using as few as 5 reference images yields a comparable accuracy of approximately 84.5%, while a single reference leads to noticeable instability due to individual sample variance. Therefore, we set *M*=10 as a conservative and reliable choice, and do not include the cases of *M*=1 and *M*=5 in the figure. This insensitivity demonstrates that even a very small set of authentic samples is sufficient to reliably estimate the real-class center, highlighting the practical deployability of our training-free scheme.

### D. Robustness Analysis

To evaluate the robustness of the proposed method against common image degradations, we conduct experiments on two representative generative models: StarGAN (a GAN-based model) and Midjourney (a diffusion-based model). The results under JPEG compression and resizing are summarized in Fig. 5.

**JPEG compression.** Left part of Fig. 5 reports the detection accuracy under different JPEG quality factors. For StarGAN, the accuracy decreases from 0.9695 (clean) to 0.924 at QF=30, representing a decline of 4.55 percentage points. For Midjourney, the accuracy drops from 1.000 (clean) to 0.951 at QF=30, representing a decline of 4.9 percentage points. As the quality factor increases from 30 to 100, the accuracy of both models recovers progressively, approaching their respective clean baselines at QF=100. Notably, the performance drop is more pronounced for Midjourney under heavy compression (QF=30 and 50), which may be attributed to the fact that diffusion-generated traces are predominantly encoded in high-frequency components that are more vulnerable to JPEG quantization. Despite these observable degradations, both models maintain accuracy well above 0.92 under the most severe compression, substantially exceeding the random guessing baseline of 0.500.

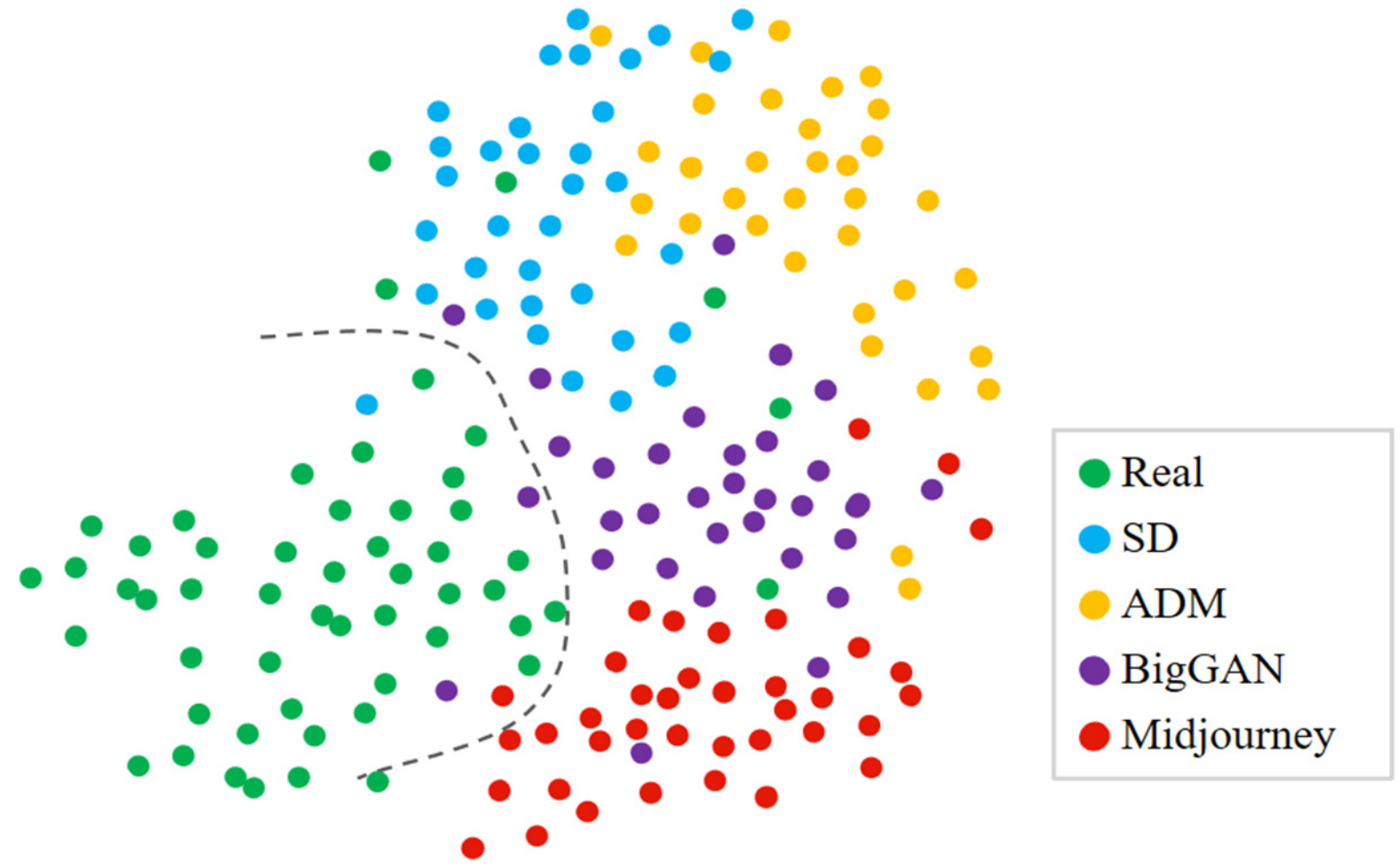

Fig. 6. t-SNE visualization of the final fused features for real and multiple AI-generated images.

**Resizing.** Right part of Fig. 5 illustrates the impact of image scaling. For StarGAN, the accuracy ranges from 0.931 at scale=0.5 to 0.9695 at scale=1.0, with a maximum deviation of 3.85 percentage points from its clean baseline. For Midjourney, the accuracy ranges from 0.958 at scale=0.5 to 1.000 at scale=1.0, with a maximum deviation of 4.2 percentage points. Both models exhibit the lowest accuracy when the image is severely downscaled to 0.5×, as substantial information loss occurs in the spatial domain. Notably, upscaling (scale=1.5 and 2.0) also leads to moderate performance degradation, likely because interpolation operations introduce smoothing effects and synthetic artifacts that interfere with noise residual extraction. Nevertheless, the ACC values keep above 0.93 across all scales, demonstrating that the proposed method retains sufficient discriminative power under resolution modifications.

Overall, the proposed method exhibits reasonable robustness to both JPEG compression and resizing under various degradation levels. While the detection accuracy degrades to some extent under heavy degradations, the performance consistently stays well across all tested adverse scenarios, particularly for diffusion-based images. This confirms that the noise-residual features, although not entirely immune to common image transformations, maintain adequate forensic discriminability for practical deployment.

### *E. Feature Visualization*

To qualitatively assess feature discriminability, t-SNE visualization of the final fused features is shown in Fig. 6. Real images are predominantly concentrated in the left region, forming a compact cluster clearly separated from all AI-generated clusters by a dashed grey line. In contrast, generated images from different models collectively occupy the right region, with no significant overlap with the real cluster. This clear separation indicates that the fused features, which combine noise-level statistics with semantic representations extracted by the ViT, possess strong discriminative power for distinguishing AI-generated images from real ones.

## IV. CONCLUSION

In this paper, a training-free detection scheme for AI-generated images is proposed by exploiting intrinsic noise-level discrepancies between real and synthetic images. A pre-trained Noiseprint++ model extracts noise residual fingerprints, and a Vision Transformer captures multi-scale features that are adaptively fused with handcrafted statistical features. An unsupervised clustering strategy guided by a small set of real reference images distinguishes real and generated images without any model training. Experiments on four benchmarks show that our method yields state-of-the-art average accuracy and strong cross-model generalization, especially on diffusion-generated images. Ablation and robustness studies verify the efficacy of each module and its resilience to common image degradations. How to further enhance the generalization to early GAN-generated images and extend the framework to video forgery detection will be investigated in future work.

## REFERENCES

[1] Karras T, Aila T, Laine S, et al. “Progressive growing of GANs for improved quality, stability, and variation,” in *Proc. Int. Conf. on Learn. Representations*, 2018.

[2] Karras T, Laine S, Aila T. “A style-based generator architecture for generative adversarial networks,” in *Proc. IEEE/CVF Conf. Comput. Vis. Pattern Recognit.* 2019, pp. 4401-4410.

[3] Choi Y, Choi M, Kim M, et al. “StarGAN: Unified generative adversarial networks for multi-domain image-to-image translation,” in *Proc. IEEE Conf. Comput. Vis. Pattern Recognit.* 2018, pp. 8789-8797.

[4] Rombach R, Blattmann A, Lorenz D, et al. “High-resolution image synthesis with latent diffusion models,” in *Proc. IEEE/CVF Conf. Comput. Vis. Pattern Recognit.* 2022, pp. 10684-10695.

[5] Ramesh A, Dhariwal P, Nichol A, et al. “Hierarchical text-conditional image generation with CLIP latents,” *arXiv:2204.06125*, 2022.

[6] "Midjourney," 2022. [Online]. Available: https://www.midjourney.com.

[7] Marra F, Gragnaniello D, Cozzolino D, et al. “Detection of GAN-generated fake images over social networks,” in *Proc. IEEE Conf. Multimedia Inf. Process. Retrieval.* 2018, pp. 384-389.

[8] Gragnaniello D, Cozzolino D, Marra F, et al. “Are GAN generated images easy to detect? A critical analysis of the state-of-the-art,” in *Proc. IEEE Int. Conf. Multimedia Expo.* 2021, pp. 1-6.

[9] Durall R, Keuper M, Keuper J. “Watch your up-convolution: CNN based generative deep neural networks are failing to reproduce spectral distributions,” in *Proc. IEEE/CVF Conf. Comput. Vis. Pattern Recognit.* 2020, pp. 7890-7899.

[10] Frank J, Eisenhofer T, Schönherr L, et al. “Leveraging frequency analysis for deep fake image recognition,” in *Proc. Int. Conf. Mach. Learn.* 2020, pp. 3247-3258.

[11] Corvi R, Cozzolino D, Zingarini G, et al. “On the detection of synthetic images generated by diffusion models,” in *Proc. IEEE Int. Conf. on Acoustics, Speech and Signal Processing.* 2023, pp. 1-5.

[12] Wang S Y, Wang O, Zhang R, et al. “CNN-generated images are surprisingly easy to spot... for now,” in *Proc. IEEE/CVF Conf. Comput. Vis. Pattern Recognit.* 2020, pp. 8695-8704.

[13] Ojha U, Li Y, Lee Y J. “Towards universal fake image detectors that generalize across generative models,” in *Proc. IEEE/CVF Conf. Comput. Vis. Pattern Recognit.* 2023, pp. 24480-24489.

[14]Radford A, Kim J W, Hallacy C, et al. "Learning transferable visual models from natural language supervision," in *Proc. Int. Conf. Mach. Learn.* 2021, pp. 8748-8763.
[15]Cozzolino D, Poggi G, Corvi R, et al. "Raising the bar of AI-generated image detection with CLIP," in *Proc. IEEE/CVF Conf. Comput. Vis. Pattern Recognit. Workshops.* 2024, pp. 4356-4366.
[16]Tao R, Wang Z, Li Y, et al. "C2P-CLIP: Injecting category common prompt in CLIP to enhance generalization in deepfake detection," in *Proc. AAAI Conf. Artif. Intell.* 2025, pp. 5261-5269.
[17]Wang Z, Bao J, Zhou W, et al. "DIRE for diffusion-generated image detection," in *Proc. IEEE/CVF Int. Conf. Comput. Vis.* 2023, pp. 22445-22455.
[18]Chen B, Zeng J, Yang J, et al. "DRCT: Diffusion reconstruction contrastive training towards universal detection of diffusion generated images," in *Proc. Int. Conf. Mach. Learn.* 2024, pp. 7621-7639.
[19]Tan C, Zhao Y, Wei S, et al. "Learning on gradients: Generalized artifacts representation for GAN-generated images detection," in *Proc. IEEE/CVF Conf. Comput. Vis. Pattern Recognit.* 2023, pp. 12105-12114.
[20]Yan S, Li O, Cai J, et al. "A sanity check for AI-generated image detection," in *Proc. Int. Conf. Learn. Represent.* 2025, pp. 70702-70720.
[21]Li O, Cai J, Hao Y, et al. "Improving synthetic image detection towards generalization: An image transformation perspective," in *Proc. ACM SIGKDD Conf. Knowledge Discovery Data Mining.* 2025, pp. 2405-2414.
[22]Ricker J, Lukovnikov D, Fischer A. "AEROBLADE: Training-free detection of latent diffusion images using autoencoder reconstruction error," in *Proc. IEEE/CVF Conf. Comput. Vis. Pattern Recognit.* 2024, pp. 9130-9140.
[23]He Z, Chen P Y, Ho T Y. "RIGID: A training-free and model-agnostic framework for robust AI-generated image detection," *arXiv preprint arXiv:2405.20112*, 2024.
[24]Yao S, Tao R, Zheng X, et al. "Leveraging failed samples: A few-shot and training-free framework for generalized deepfake detection," in *Proc. AAAI Conf. Artif. Intell.* 2026, pp. 27764-27772.
[25]Bai J, Lin M, Cao G, Lou Z. "AI-generated video detection via spatial-temporal anomaly learning," in *Proc. Chinese Conf. on Pattern Recog. Computer Vision.* 2024, pp. 460-470.
[26]Guillaro F, Cozzolino D, Sud A, et al. "TruFor: Leveraging all-round clues for trustworthy image forgery detection and localization," in *Proc. IEEE/CVF Conf. Comput. Vis. Pattern Recognit.* 2023, pp. 20606-20615.
[27]Lou Z, Cao G, Lin M, et al. "Trusted video inpainting localization via deep attentive noise learning," in *IEEE Trans. on Dependable and Secure Computing.* 2025, pp. 7215-7228.
[28]Guo K, Zhu H, Cao G. "Effective image tampering localization via enhanced transformer and co-attention fusion," in *Proc. IEEE Int. Conf. on Acoustics, Speech and Signal Processing.* 2024, pp. 4895-4899.
[29]Zhao M, Cao G, Huang X, Yang L. "Hybrid transformer-CNN for real image denoising," in *IEEE Signal Process. Letters.* 2022, pp. 1252-1256.
[30]Dosovitskiy A, Beyer L, Kolesnikov A, et al. "An image is worth 16x16 words: Transformers for image recognition at scale," *arXiv preprint arXiv:2010.11929*, 2020.
[31]Lou Z, Cao G, Guo K, et al. "Exploring multi-view pixel contrast for general and robust image forgery localization," in *IEEE Trans. on Info. Forensics and Security*. 2025, pp. 2329-2341.
[32]Zhu M, Chen H, Yan Q, et al. "GenImage: A million-scale benchmark for detecting AI-generated image," in *Proc. Adv. Neural Inf. Process. Syst.* 2023, pp. 77771-77782.
[33]Bammey Q. "Synthbuster: Towards detection of diffusion model generated images," in *IEEE Open J. Signal Process.* 2023, pp. 1-9.
[34]Zhong N, Xu Y, Li S, et al. "PatchCraft: Exploring texture patch for efficient AI-generated image detection," *arXiv preprint arXiv:2311.12397*, 2023.

[35]Karras T, Laine S, Aittala M, et al. "Analyzing and improving the image quality of stylegan." in *Proc. IEEE/CVF Conf. Comput. Vis. Pattern Recognit.* 2020, pp. 8110-8119.

[36]Brock A, Donahue J, Simonyan K. "Large scale GAN training for high fidelity natural image synthesis." *arXiv preprint arXiv:1809.11096*, 2018.

[37]Xu Z, Karaman S, Chang S-F. "Detecting and simulating artifacts in gan fake images." in *Proc. IEEE Int. Workshop Info. Forensics and Security.* 2019, pp. 1-6.